\documentclass[runningheads]{llncs}

\usepackage{eccv}

\usepackage{eccvabbrv}
\usepackage{graphicx}
\usepackage{booktabs}
\usepackage{multirow}
\usepackage{multicol}
\usepackage{xcolor}         
\usepackage{amsmath}
\usepackage[ruled,linesnumbered]{algorithm2e}

\usepackage[accsupp]{axessibility}  

\usepackage{hyperref}

\usepackage{orcidlink}

\begin{document}


\makeatletter
\def\@fnsymbol#1{\ensuremath{%
  \ifcase#1%
  \or\dagger%
  \or\star%
  \or\star\star%
  \or\star\star\star%
  \or\ddagger%
  \or\mathchar"278%
  \or\mathchar"27B%
  \or\|%
  \or **%
  \or\dagger\dagger%
  \or\ddagger\ddagger%
  \else\@ctrerr%
  \fi}}
\makeatother

\title{Purify then Guide: Rethinking Domain Generalization for Multimodal Face Anti-Spoofing} 

\titlerunning{Purify then Guide}

\author{
Yingjie Ma\inst{1,2}\thanks{These authors contributed equally to this work.}
\and
Xun Lin\inst{2,8}\protect\footnotemark[1]
\and
Zitong Yu\inst{2,3}\thanks{Corresponding authors}
\and
Haonan Wang\inst{5}
\and
Ruixin Zhang\inst{5}\protect\footnotemark[2]
\and
Shouhong Ding\inst{5}
\and
Xin Liu\inst{6}
\and
Xiaochen Yuan\inst{7}
\and
Weicheng Xie\inst{1,4}
\and
Linlin Shen\inst{1,4}\protect\footnotemark[2]
}

\authorrunning{Y.~Ma et al.}

\institute{
College of Computer Science and Software Engineering, Shenzhen University
\and
School of Computing and Information Technology, Great Bay University
\and
Dongguan Key Laboratory for Intelligence and Information Technology
\and
Guangdong Provincial Key Laboratory of Intelligent Information Processing, Shenzhen University
\and
Tencent Youtu Lab
\and
School of Psychology, Shanghai Jiao Tong University
\and
Faculty of Applied Sciences, Macao Polytechnic University
\and
The Chinese University of Hong Kong
\and
\url{https://github.com/murInJ/MMDA}
}

\maketitle

\vspace{-2.5em}
\begin{abstract}
    Face anti-spoofing (FAS) is critical for secure deployment of face recognition in high-stakes applications. However, existing multimodal FAS methods still generalize poorly to unseen domains. We argue that a key reason is that they directly align noisy multimodal features in which spoof-relevant cues are entangled with modality- and domain-specific variations, so both useful and nuisance patterns are forced to match across domains. 
    Motivated by this, we design \textbf{M}ultimodal \textbf{D}enoising and \textbf{A}lignment (\textbf{MMDA}), a CLIP-based  framework that (i) purifies multimodal features, (ii) softly guides them into a semantic space, and (iii) preserves the pre-trained geometry during task adaptation. To decouple spoof cues from these variations before alignment, the \textbf{M}odality-\textbf{D}omain Joint \textbf{D}ifferential \textbf{A}ttention (\textbf{MD2A}) module contrasts same-domain cross-sample features to down-weight patterns shared within each domain/modality, which helps reduce domain- and modality-related biases while better preserving discriminative spoof information. To avoid destructive, over-rigid alignment, the \textbf{R}epresentation \textbf{S}pace \textbf{S}oft (\textbf{RS2}) strategy aligns visual features to real/spoof subspaces spanned by multiple prompts in the CLIP space, providing a soft semantic pull instead of collapsing features onto a single text embedding. To prevent task-specific fine-tuning from destroying CLIP's generalizable structure, the \textbf{U}-shaped \textbf{D}ual \textbf{S}pace \textbf{A}daptation (\textbf{U-DSA}) module introduces deep yet parameter-efficient adaptation and remaps adapted features back to shallower, more domain-invariant layers. 
    Extensive experiments on WMCA, CeFA, PADISI, and SURF under complete-modality, missing-modality, and limited-source protocols show that MMDA consistently surpasses state-of-the-art methods, reducing average HTER by up to 9.63\% and improving AUC by up to 5.98\% in cross-domain evaluation.
  \vspace{-1.0em} 
\end{abstract}

\section{Introduction}
\label{sec:intro}
Facial recognition (FR) systems are widely deployed in high-stakes applications such as payment, access control, and surveillance~\cite{yu2022deep,xu2023multimodal}, and thus require reliable face anti-spoofing (FAS)~\cite{cai2024s,yu2024benchmarking,jiang2024open}. Modern FAS systems increasingly adopt multimodal sensing with RGB, infrared (IR), and DEPTH streams, since multimodal inputs provide complementary cues on appearance, geometry, and material properties that are crucial for distinguishing real faces from spoof artifacts. However, multimodal FAS still generalizes poorly to unseen domains: performance often drops sharply under new cameras, environments, and attack types.

\begin{figure}[t]
\centering
\includegraphics[width=0.90\textwidth]{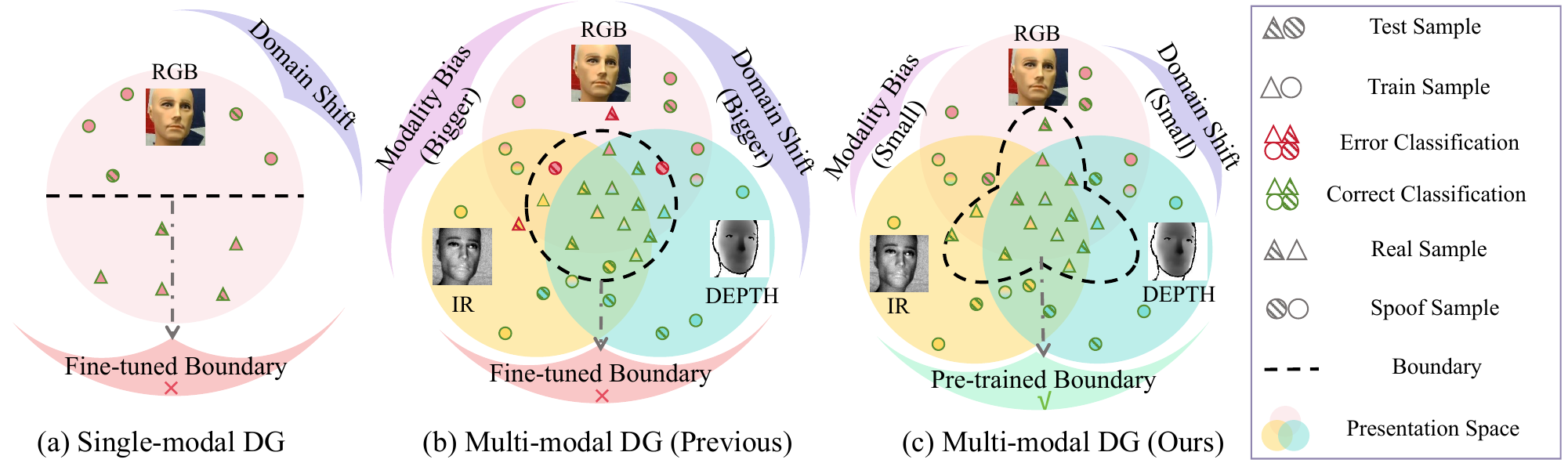}
\vspace{-1.0em}
\caption{
(a) In single-modal FAS, domain shifts between training and testing environments induce domain gaps that degrade generalization. (b) In multimodal FAS, existing methods typically fine-tune a backbone and classifier on all source domains to learn a new decision boundary in the joint multimodal space; this boundary is often overly smoothed to accommodate diverse domain--modality combinations, causing hard target samples to lie near the boundary. (c) Our MMDA framework instead first reduces domain and modality gaps in the representation and then aligns multimodal features to a well-structured real/spoof separation inherited from the pre-trained space, improving separability under severe shifts. \textbf{Note:} Face images are from the WMCA~\cite{george2019biometric} dataset.}
\label{fig1}
\vspace{-2.0em}
\end{figure}

A distinctive difficulty of multimodal cross-domain FAS (beyond single modality DG FAS) is that generalization is challenged by two coupled shifts.
First, the fused representation can drift across groups because cross-modal relations (e.g., complementarity, reliability, and scale among RGB/IR/DEPTH) are themselves domain dependent.
Second, learning a new decision boundary by fine-tuning on limited source domains often yields an overly smoothed separator in the joint multimodal space, leaving hard target samples close to the boundary (Fig.~\ref{fig1}(b)).
To better understand the representation drift, we conduct two simple diagnostics on per-class embeddings for each modality (and the fused representation) and observe consistent cross-domain discrepancies.
(i) An ANOVA-style ratio $r=\mathrm{Var}_{\mathrm{between}}/\mathrm{Var}_{\mathrm{total}}$ indicates that cross-domain variation, especially for bonafide (real) samples, exhibits a strong mean-shift component, suggesting the presence of an approximately additive bias.
(ii) Same-domain pairs are significantly closer than cross-domain pairs under both $\ell_2$ and cosine distances with separated bootstrap confidence intervals, implying that local neighborhoods are relatively stable within a group.
Taken together, these statistics suggest that a non-trivial portion of the cross-group discrepancy behaves like a group-shared residual that can be estimated under same-group conditioning and removed before any alignment.

These observations motivate our \textit{purify first, then guide} design.
Instead of directly aligning noisy multimodal features, where spoof-relevant cues are entangled with domain/modality artifacts, we first \textit{purify} the fused embedding by suppressing group-dependent components, and then \textit{guide} the purified features into a stable real/spoof geometry that resists boundary drift.
To this end, we view a fused feature as the sum of invariant spoof cues and variant artifacts. To operationalize this view at the fusion level, we adopt a denoising-style attention strategy and reformulate differential attention for the multimodal DG setting. Conventional Differential Attention (DA)~\cite{ye2024differential} applies two attention branches to views of the same sample and subtracts their responses to suppress sample-wise spurious patterns. In our \textbf{M}odality-\textbf{D}omain Joint \textbf{D}ifferential \textbf{A}ttention (\textbf{MD2A}) module, each sample keeps its own multimodal features in the main attention branch, while the noise branch is fed with features from another sample in the same domain. The two branches thus share domain conditions and modality configuration but differ in identity and attack content. The differential response between them predominantly captures domain- and modality-specific components rather than semantics that distinguish real from spoof. Treating this response as a residual and subtracting it from the main branch output can help reduce shared domain-/modality-related variations, allowing the fused multimodal representation to better emphasize spoof-discriminative cues.

Once the features are purified, the next question is how to align them without destroying their structure. Aligning to a target defined only by limited source-domain data risks encoding the same biases we aim to remove, and hard alignment of visual features to a single text embedding can collapse the rich spoof manifold. To avoid such destructive alignment, we leverage a pre-trained CLIP model and its joint visual--text space as a strong, domain-agnostic anchor for real/spoof separation. We encode multiple textual descriptions of real and spoof faces into a text-defined subspace and introduce a \textbf{R}epresentation \textbf{S}pace \textbf{S}oft (\textbf{RS2}) alignment scheme that softly pulls multimodal visual features toward appropriate regions of this subspace. RS2 imposes a soft semantic constraint that guiding features toward the correct semantic neighborhood instead of pinning them to a single prompt, so that the model can exploit CLIP’s semantic structure without losing fine-grained spoof cues.

Finally, we address a deeper tension in using a pre-trained model for FAS: without adapting deep layers, the feature extractor lacks task-specific capacity; yet aggressively fine-tuning deep layers tends to distort the pre-trained representation space and harm the generalization of its decision boundary. To balance these effects, we design a \textbf{U}-shaped \textbf{D}ual \textbf{S}pace \textbf{A}daptation (\textbf{U-DSA}) module. U-DSA introduces a deep, parameter-efficient adaptation pathway on top of CLIP while keeping the original backbone space as a stable anchor. The adapted deep features are then mapped back to shallower, more domain-invariant layers through a U-shaped structure, and RS2 is applied at multiple depths. This builds a bridge between a task-specialized space and the pre-trained space, enabling the model to increase task capacity while preserving the generalizable CLIP geometry after MD2A has reduced domain and modality gaps.
To sum up, our contributions are five-fold:
\begin{itemize}
\item We formulate multimodal DG FAS as a ``purify then guide'' problem and propose MMDA, a CLIP-based framework that first purifies multimodal features and then performs soft, non-destructive alignment.
\vspace{0.5mm}
\item We introduce MD2A, which reformulates differential attention by feeding same-domain different-sample features into the noise branch, explicitly attenuating domain- and modality-specific artifacts in multimodal fusion while preserving discriminative spoof information. Moreover, we theoretically show that the proposed MD2A can tighten the generalization error bound in multimodal FAS scenarios.
\vspace{0.5mm}
\item We develop RS2 and U-DSA, which together perform global and layer-wise soft alignment to a CLIP-based real/spoof representation space, balancing task-specific adaptation and preservation of pre-trained geometry.
\vspace{0.5mm}
\item We conduct extensive experiments on four multimodal FAS benchmarks under complete-modality, missing-modality, and limited-source protocols, achieving state-of-the-art cross-domain performance.
\end{itemize}

\section{Related Work}
\label{sec:related}

\subsection{Face Anti-Spoofing}

Deep learning has driven rapid progress in face anti-spoofing (FAS) \cite{shi2025shield,wang2026faceshield,wang2026micro}, with numerous architectures designed to extract discriminative cues for separating real from spoof faces. However, performance typically degrades severely under domain shifts (e.g., changes in lighting, cameras, or acquisition setups)~\cite{yu2022deep,huang2022adaptive,yu2025multi,li2025optimal,lin2025instructflip}. To improve domain generalization (DG), recent works explore adversarial learning~\cite{jiang2023adversarial,yue2023cyclically}, feature disentanglement~\cite{liu2024quality,jiang2024open,zhou2023instance}, meta-learning~\cite{cai2022learning,du2022energy}, data augmentation~\cite{cai2024towards,ge2025difffas}, and domain alignment~\cite{hu2024rethinking,wang2024csdg,le2024gradient,sun2023rethinking}, aiming to learn domain-invariant feature representations and more robust decision boundaries. These DG-oriented methods are almost exclusively developed for unimodal FAS and do not account for multimodal fusion or modality-specific biases.

Multimodal FAS integrates multiple sensing streams to leverage complementary appearance, geometric, and material cues~\cite{liu2023fm,kong2022beyond,kong2024m,yu2020multi,xie2024fusionmamba,ma2026pa,lin2025learning,lin2025reliable,zhang2026multimodal,ma2026cm}. Attention-based fusion and adaptive loss functions have been used to extract complementary information~\cite{george2021cross,zheng2025towards}, while cross-modal translation is explored to reduce semantic discrepancies across modalities~\cite{liu2021face}. Recent works further study robustness under missing-modality conditions by proposing new protocols and methods for flexible multimodal FAS~\cite{yu2023visual,yu2024rethinking,yu2023flexible}. To support arbitrary modality combinations, cross-modal attention and multimodal adapters with pre-trained ViTs are used to learn modality-insensitive features~\cite{liu2023ma,liu2023fm,lin2024suppress}. Among multimodal FAS methods that explicitly consider DG, MMDG~\cite{lin2024suppress} is a representative work: it analyzes the conflict between DG objectives and multimodal fusion, and introduces uncertainty rectification and modality re-balancing to alleviate this issue. In this paper, we follow this line of work and further investigate representation-space methods that jointly handle domain and modality gaps for multimodal DG FAS using a CLIP-based framework.

\subsection{CLIP-based FAS}

Parameter-efficient transfer learning adapts large pre-trained models such as Vision Transformers (ViT)~\cite{dosovitskiy2020image} and CLIP~\cite{radford2021learning} to new tasks by tuning a small subset of parameters, improving efficiency and reducing overfitting. PETL-style designs have shown strong performance in FAS~\cite{cai2023rehearsal,cai2024s,srivatsan2023flip}. For example, S-Adapter~\cite{cai2024s} inserts lightweight adapter modules into pre-trained backbones to adjust features in a parameter-efficient way, and rehearsal-based PETL frameworks~\cite{cai2023rehearsal} and FLIP-like designs~\cite{srivatsan2023flip} further demonstrate the benefit of adapting only a small number of parameters for robust face anti-spoofing performance.

Leveraging CLIP as a backbone, several works explore CLIP-based and vision–language FAS models~\cite{liu2024cfpl,srivatsan2023flip,fang2024vl,liu2025bottom}, typically using text prompts to provide semantic guidance and improve generalization~\cite{gao2024clip}. These methods, however, are mostly unimodal or rely on simple late fusion, and primarily focus on how to update or regularize pre-trained weights and prompts. In multimodal DG FAS, the structure of the pre-trained representation space itself may offer further potential to be exploited or preserved, which motivates exploring representation-space-oriented designs on top of CLIP.

\section{Method}

\begin{figure*}[t]
\centering
\includegraphics[width=0.90\textwidth]{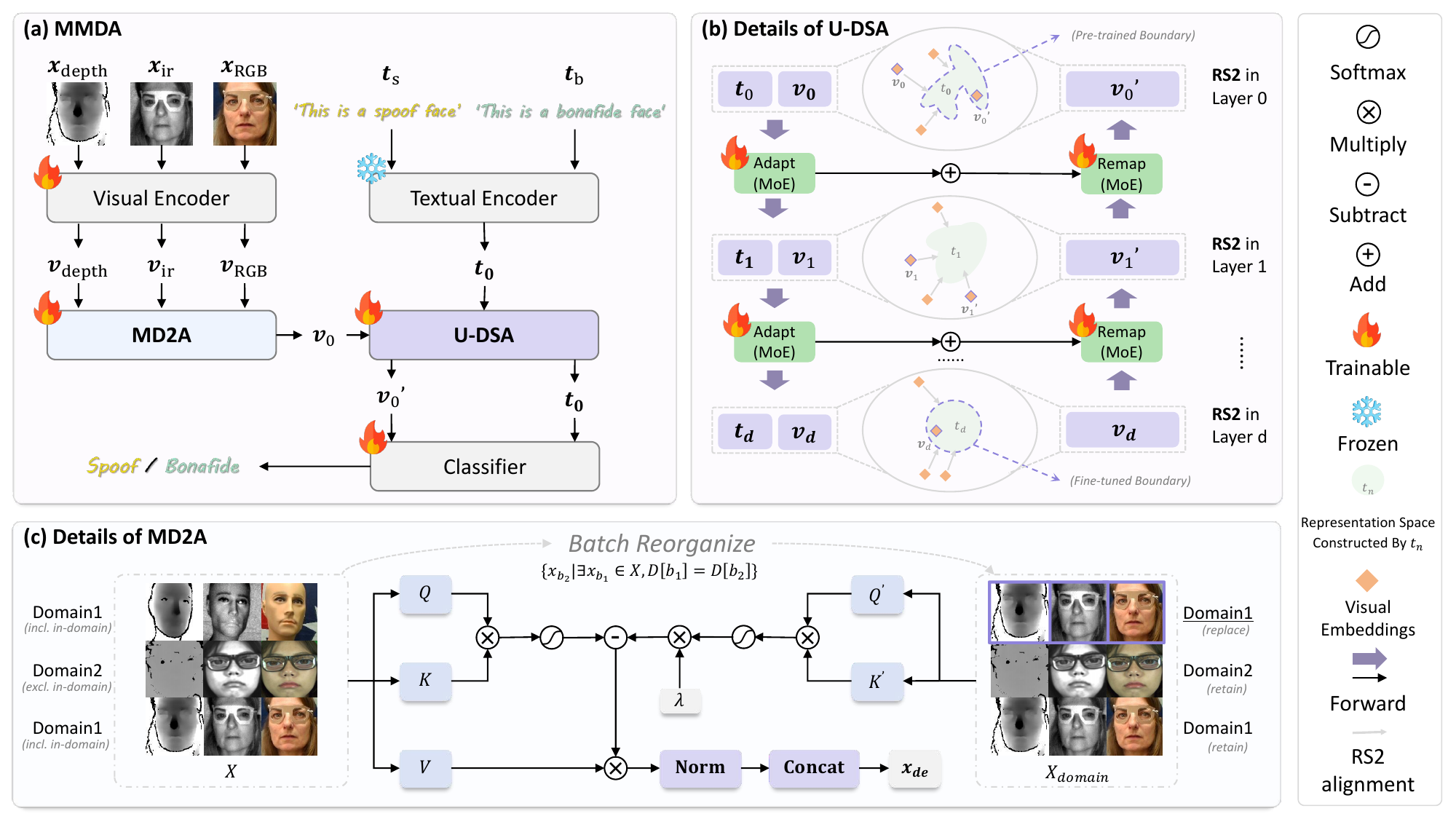}
\vspace{-0.2em}
\caption{
Overall framework of MMDA. (a) Following the ``purify then guide'' principle, MMDA first denoises multimodal visual embeddings with Modality-Domain Joint Differential Attention (MD2A), and then performs soft, non-destructive alignment in a CLIP-based representation space. (b) Details of the U-shaped Dual Space Adaptation (U-DSA) module and the layer-wise use of the Representation Space Soft (RS2) alignment scheme, which together balance task-specific adaptation and preservation of CLIP geometry. (c) Operational details of MD2A, where cross-sample same-domain pairing is used to estimate domain/modality-shared residuals. \textbf{Note:} All face images shown are from the WMCA~\cite{george2019biometric}, CeFA~\cite{liu2021casia}, PADISI~\cite{rostami2021detection}, and SURF~\cite{zhang2020casia} datasets.}
\label{U-DSA}
\vspace{-1.0em}
\end{figure*}

    
    
    


\begin{figure}[t]
  \centering
  \begin{minipage}{1.0\textwidth}
    \begin{algorithm}[H]
    \caption{Modality--Domain Joint Differential Attention (MD2A)}
    \label{Algorithm:1}
    \footnotesize
    \SetInd{0.2em}{0.2em}
    \KwIn{batch multimodal features $X = \{ \boldsymbol x_i \}_{i=1}^B$; domain labels $\mathcal D = \{\boldsymbol d_i\}_{i=1}^B$}
    \KwOut{purified multimodal features $X_{de}$}
    
    \textcolor{teal}{\small \# 1. Construct same-domain pairs (preferably different samples)}\\
    \For{$i \gets 1$ \KwTo $B$}{
        choose $j$ s.t. $\mathcal D[j] = \mathcal D[i]$ (and $j \neq i$ if possible)\;
        $\tilde{\boldsymbol x}_i \gets \texttt{Concat}(\boldsymbol x_i, \boldsymbol x_j)$\;
    }
    $\tilde X \gets \{\tilde{\boldsymbol x}_i\}_{i=1}^B$\;
    
    \textcolor{teal}{\small \# 2. Project and split into main / differential branches}\\
    $\mathit Q_{\text{all}} \gets \tilde X W_q,\quad \mathit K_{\text{all}} \gets \tilde X W_k,\quad \mathit V \gets \tilde X W_v$\;
    $(\mathit Q, \mathit Q') \gets \texttt{split}(\mathit Q_{\text{all}})$,\quad $(\mathit K, \mathit K') \gets \texttt{split}(\mathit K_{\text{all}})$\;
    
    \textcolor{teal}{\small \# 3. Differential attention for feature purification}\\
    $s \gets 1/\sqrt{n_d}$\;
    $\mathit A \gets \mathit Q \mathit K^\top \cdot s$,\quad $\mathit A' \gets \mathit Q' \mathit K^{\prime\top} \cdot s$\;
    $X_{de} \gets \big(\texttt{softmax}(\mathit A) - \lambda \texttt{softmax}(\mathit A')\big)\mathit V$\;

    \end{algorithm}
  \end{minipage}
  \vspace{-2.0em}
\end{figure}

Our MMDA framework is illustrated in Fig.~\ref{U-DSA}(a). Following the ``purify then guide'' principle, we first reduce domain and modality artifacts in multimodal visual embeddings, and then softly align the purified features in a CLIP-based representation space. Concretely, input RGB/IR/DEPTH images and text prompts are fed into the CLIP image and text encoders to obtain visual and textual embeddings. Throughout training, we keep the CLIP text encoder frozen and fine-tune the CLIP image encoder on the FAS data. The resulting multimodal visual embeddings are first fused and denoised by Modality-Domain Joint Differential Attention (MD2A), and are then passed through the proposed U-shaped Dual Space Adaptation (U-DSA) module and aligned to text-defined real/spoof subspaces via the Representation Space Soft (RS2) scheme. Finally, a shallow classifier operates jointly on the adapted visual embeddings and text embeddings for real/spoof prediction.

\vspace{-0.8em}
\subsection{Preliminary and Motivation}
\vspace{-0.6em}
\noindent\textbf{CLIP for FAS and a hidden risk.}
A common CLIP-based FAS recipe~\cite{srivatsan2023flip,lin2025instructflip} is to construct text prompts describing real/spoof faces and classify an image by its similarity to these prompts, optionally combined with parameter-efficient tuning of the visual branch.
However, the CLIP pre-training objective does not explicitly enforce cross-domain or cross-modality consistency for FAS.
In multimodal DG FAS, directly fine-tuning the visual encoder and learning a new decision boundary from limited source domains can therefore (i) overfit to acquisition-specific artifacts and (ii) distort the pre-trained geometry, i.e., the decision boundary drifts away from a more transferable separator induced by the frozen text space.
This tension motivates using CLIP as an anchor while constraining task adaptation to be non-destructive.

\vspace{0.5mm}
\noindent\textbf{Groups: domain and modality setting.}
Each multimodal sample is collected under a domain $d\in\mathcal{D}$ (dataset/camera/environment) and a modality setting $m\in\mathcal{M}$ (available sensing streams, e.g., RGB+IR+DEPTH or missing IR/DEPTH).
We denote a group by $g=(d,m)\in\mathcal{G}:=\mathcal{D}\times\mathcal{M}$.
Different groups induce different acquisition biases and fusion behaviors, so multimodal embeddings can shift systematically across $g$.

\vspace{0.5mm}
\noindent\textbf{A decomposition view and the ``purify then guide'' principle.}
Let $h(x)$ denote the fused multimodal embedding before our purification module.
Motivated by the empirical mean-shift evidence, we adopt a simple decomposition view:
\begin{equation}
h(x)= s(x,y) + \Delta_g + \varepsilon,
\end{equation}
where $s(x,y)$ captures spoof-relevant, group-invariant semantics and $\Delta_g$ is a group-dependent bias term.
This suggests two design requirements for multimodal DG FAS:
(i) \textit{Purify}: suppress $\Delta_g$ before any alignment, so that cross-group discrepancies (the ``shift term'' in our analysis) are reduced.
(ii) \textit{Guide without destroying geometry}: align purified features to a stable semantic structure while preventing boundary drift, which corresponds to constraining the learned separator to stay close to a pre-trained one.

Concretely, our MD2A operationalizes (i) by using same-group conditioning (in practice, same-domain pairing; and same modality setting when applicable) to estimate group-shared residual patterns and subtract them, producing a purified embedding that is easier to align robustly.
Based on purified features, RS2 and U-DSA address (ii): RS2 provides a soft semantic pull by aligning to class-specific text subspaces spanned by multiple prompts, and U-DSA increases task capacity via deep adaptation while remapping back to shallower, more domain-invariant spaces to preserve CLIP geometry.

\vspace{-0.5em}
\subsection{Modality-Domain Joint Differential Attention}
\vspace{-0.3em}
MD2A implements the \textit{purify} step in our framework by explicitly estimating and suppressing domain- and modality-specific residuals in multimodal features. As shown in Fig.~\ref{U-DSA}(c), let $X = \{\boldsymbol x_0,\dots,\boldsymbol x_b\}$ denote fused multimodal features in a mini-batch and $\mathcal D$ the corresponding domain labels. For each sample $\boldsymbol x_i$, we randomly select another sample $\tilde{\boldsymbol x}_i$ from the same domain:
\begin{equation}
\small
\mathcal D[\tilde{\boldsymbol x}_i] = \mathcal D[\boldsymbol x_i], \quad
\tilde{\boldsymbol x}_i \neq \boldsymbol x_i \ \text{if available},
\end{equation}
and fall back to $\tilde{\boldsymbol x}_i = \boldsymbol x_i$ when no other same-domain sample exists. In practice, we implement this pairing by reorganizing the batch as in Alg.~\ref{Algorithm:1}, where each row of $X$ is replaced by the feature-wise concatenation of $(\boldsymbol x_i, \tilde{\boldsymbol x}_i)$ using \texttt{Concat}, which is applied in both training and inference.

We then project the concatenated features and split them into a main branch and a differential branch:
\begin{equation}
\small
\begin{aligned}
(\mathit Q, \mathit Q') = \text{split}(X W_q), (\mathit K, \mathit K') = \text{split}(X W_k), \mathit V = X W_v,
\end{aligned}
\end{equation}
where $W_q, W_k, W_v$ are learnable projection matrices and $n_d$ is the feature dimension. The attention maps of the two branches are
\begin{equation}
\small
\mathit A = \frac{\mathit Q \mathit K^\top}{\sqrt{n_d}}, \quad
\mathit A' = \frac{\mathit Q' \mathit K^{\prime\top}}{\sqrt{n_d}}.
\end{equation}
The differential branch learns a group-shared residual bias associated with $\Delta_g$ and suppresses it in the fused features. The MD2A output is obtained by contrasting the two branches:
\begin{equation}
\small
X_{de} =
\left[ \text{softmax}(\mathit A) - \lambda \, \text{softmax}(\mathit A') \right] \mathit V,
\end{equation}
where $\lambda$ controls the strength of the differential branch, and we follow the original differential attention~\cite{ye2024differential} for its parameterization.
\begin{equation}
\small
\lambda=\text{exp}(\lambda_{q1}\cdot\lambda_{k1})-\text{exp}(\lambda_{q2}\cdot\lambda_{k2})+0.8-0.6\times \text{exp}(-0.3\cdot(l-1)),
\end{equation}
where $\lambda_{q1}, \lambda_{k1}, \lambda_{q2}, \lambda_{k2} \in \mathbb{R}^{n_d}$ are learnable vectors, and $l \in [1,L]$ denotes the layer index. In practice, we set $l=14$. When $\tilde{\boldsymbol x}_i = \boldsymbol x_i$ for all $i$, MD2A degenerates to the original differential attention, which suppresses sample-wise noise. By feeding same-domain (preferably different) samples into the differential branch, MD2A estimates a group-shared residual bias associated with $\Delta_g$. Subtracting this residual attenuates domain- and modality-specific artifacts while preserving sample-specific, spoof-relevant cues in the fused representation, providing a cleaner starting point for subsequent alignment.

\vspace{-0.3em}
\subsection{Representation Space Alignment}

\noindent\textbf{Representation Space Soft Alignment (RS2).}
RS2 realizes the ``guide'' step by softly pulling purified multimodal features toward appropriate regions of a text-defined real/spoof subspace in the CLIP representation space, rather than collapsing them onto a single prompt. As shown in Fig.~\ref{U-DSA}(b), given a set of captions $C$, we obtain text embeddings $T$ via the frozen CLIP text encoder. Let $V$ denote the corresponding visual embeddings produced by the CLIP image encoder, MD2A and U-DSA. RS2 encourages each visual embedding to lie close to the subspace spanned by text embeddings of the same class, while a shared classifier provides additional discriminative supervision on both modalities.

Let $\hat y_i \in \{0,1\}$ be the ground-truth label (real/spoof) of a visual embedding $\boldsymbol v_i$, and $T^{(\hat y_i)} \subset T$ the set of text embeddings for class $\hat y_i$. We define its distance to the class-specific text subspace as:
\begin{equation}
\small
d_i = \min_{\boldsymbol t_j \in T^{(\hat y_i)}} \frac{1}{2}\left(1 - \frac{\boldsymbol v_i \cdot \boldsymbol t_j}{\|\boldsymbol v_i\| \,\|\boldsymbol t_j\|}\right),
\end{equation}
i.e., the minimum cosine distance to text embeddings of the same class. To avoid over-confident targets, we apply label smoothing: for a $\epsilon=0.2$:
\begin{equation}
\small
\tilde y_i =
\begin{cases}
1 - \epsilon, & \hat y_i = 1,\\
\epsilon, & \hat y_i = 0.
\end{cases}
\end{equation}
The alignment loss is a smoothed binary cross-entropy on the distance-based score:
\begin{equation}
\small
\mathcal L_{\text{align}} = -\sum_{\boldsymbol v_i \in V} \left( 
\tilde y_i \log(1 - d_i) + (1 - \tilde y_i) \log d_i
\right).
\end{equation}

We further attach a shared binary classifier on top of both visual and text embeddings. Let $\boldsymbol e_j$ denote either a visual or a text embedding, and $p_{\boldsymbol e_j}$ the predicted probability of the real class. Using the same smoothed targets, the classification loss is:
\begin{equation}
\small
\mathcal L_{\text{cls}} = - \sum_{\boldsymbol e_j \in \{V,T\}} \left(
\tilde y_j \log p_{\boldsymbol e_j} + (1 - \tilde y_j) \log(1 - p_{\boldsymbol e_j})
\right).
\end{equation}
The RS2 loss combines both terms:
\begin{equation}
\small
\mathcal L_{\text{RS2}} = \mathcal L_{\text{cls}} + \mathcal L_{\text{align}}.
\end{equation}
Compared to hard alignment to a single text prototype, RS2 provides a soft semantic pull toward a \emph{subspace} defined by multiple prompts, allowing visual features to exploit CLIP's semantic structure while retaining fine-grained spoof cues.

\noindent\textbf{U-shaped Dual Space Adaptation (U-DSA).}
RS2 relies on a meaningful CLIP geometry, but task-specific adaptation on limited FAS data can easily distort this geometry and harm cross-domain generalization. 
Simply stacking deeper adaptation layers boosts capacity at the cost of warping the pre-trained space. 
U-DSA is designed to resolve this tension by (i) applying RS2 layer-wise and (ii) building a U-shaped dual-space pathway that feeds deep adapted features back to shallower, more domain-invariant spaces.

Let $\boldsymbol v_0$ be the input visual embedding from the CLIP image encoder (after MD2A), and let $d$ be the depth of U-DSA. The forward (bottom) path progressively applies lightweight adaptation modules:
\begin{equation}
\small
\boldsymbol v_i = \text{Adapt}_i(\boldsymbol v_{i-1}), \quad i = 1,\dots,d,
\end{equation}
where each $\text{Adapt}_i$ is implemented by a parameter-efficient MoE. 
The feedback (top) path then remaps deeper representations back to shallower spaces:
\begin{equation}
\small
\boldsymbol v'_d = \boldsymbol v_d,\quad
\boldsymbol v'_i = \boldsymbol v_i + \text{Remap}_i(\boldsymbol v'_{i+1}), \quad i = d-1,\dots,0,
\end{equation}
where $\text{Remap}_i$ (also implemented as MoE) projects the deeper representation $\boldsymbol v'_{i+1}$ into the space of layer $i$. In practice, RS2 losses are applied to intermediate representations $\boldsymbol v_i$ and $\boldsymbol v'_i$ to regularize each layer toward the text-defined subspaces.

This U-shaped dual-space design allows deeper layers to capture task-specific multimodal patterns, while the feedback path and layer-wise RS2 keep the overall representation close to the pre-trained CLIP geometry. 
As a result, the final classifier operates in a space that is both enriched by adaptation and anchored by the original real/spoof separation, completing the ``purify then guide'' pipeline after MD2A has removed most domain and modality artifacts.

\vspace{-0.2em}
\section{Experiments}
\vspace{-0.5em}
\subsection{Experimental Setup}
\vspace{-0.3em}
\noindent\textbf{Datasets and protocols.}\quad
We follow the multimodal DG protocol of MMDG~\cite{lin2024suppress} and evaluate MMDA on four benchmarks: WMCA (\textbf{W})~\cite{george2019biometric}, CeFA (\textbf{C})~\cite{liu2021casia}, PADISI (\textbf{P})~\cite{rostami2021detection}, and SURF (\textbf{S})~\cite{zhang2020casia}. 
Three types of protocols are considered: 
(i) \emph{Protocol~1} (complete modalities) uses all available modalities for training and testing under four Leave-One-Out (LOO) domain splits, e.g., \textbf{CPS$\rightarrow$W} trains on \textbf{C}, \textbf{P}, \textbf{S}, and tests on \textbf{W}; 
(ii) \emph{Protocol~2} introduces test-time missing-modality conditions on top of Protocol~1, including missing DEPTH, missing IR, and missing both DEPTH and IR; 
(iii) \emph{Protocol~3} uses only two datasets as source domains (CW$\rightarrow$PS and PS$\rightarrow$CW) to simulate limited-source training.

\noindent\textbf{Metrics.}\quad
Following prior work~\cite{lin2024suppress,yu2023visual,yu2024rethinking}, we report Half Total Error Rate (HTER) and Area Under the ROC Curve (AUC) for all protocols. 
Lower HTER and higher AUC indicate better performance.

\noindent\textbf{Implementation details.}\quad
We adopt CLIP ViT-B/16~\cite{radford2021learning} as the backbone. 
The CLIP text encoder is kept frozen, while the CLIP image encoder is fine-tuned jointly with the proposed MD2A and U-DSA modules and the final classifier on the multimodal FAS data. 
All images are resized to $224\!\times\!224$ and tokenized into $14\!\times\!14$ patch tokens with a 512-dim embedding.
We train MMDA with AdamW, a learning rate of $5\!\times\!10^{-6}$, weight decay $1\!\times\!10^{-3}$, batch size 576, and 100 epochs for all protocols. 
Unless otherwise specified, U-DSA uses 7 layers; the classifier operates on the adapted visual embeddings.

\begin{table*}[t]
 \centering
 \caption{Cross-dataset testing results under the fixed-modal scenarios (Protocol 1) among CASIA-CeFA (C), PADISI (P), CASIA-SURF (S), and WMCA (W). Best results are marked in \textbf{bold}.} 
 \vspace{-0.8em}
    \resizebox{1.00\textwidth}{!}{
    \begin{tabular}{l|cc|cc|cc|cc|cc} 
        \toprule[1.3pt]
        \multirow{2}{*}{\textbf{Method}}& \multicolumn{2}{c|}{\textbf{CPS$\rightarrow$W}}&\multicolumn{2}{c|}{\textbf{CPW$\rightarrow$S}}&\multicolumn{2}{c|}{\textbf{CSW$\rightarrow$P}}&\multicolumn{2}{c|}{\textbf{PSW$\rightarrow$C}}&\multicolumn{2}{c}{\textbf{Average}}\\
        \cmidrule(r){2-3} \cmidrule(r){4-5} \cmidrule(r){6-7} \cmidrule(r){8-9} \cmidrule(r){10-11}
        & HTER(\%)$\downarrow$ & AUC(\%)$\uparrow$ & HTER(\%)$\downarrow$ & AUC(\%)$\uparrow$& HTER(\%)$\downarrow$ & AUC(\%)$\uparrow$ & HTER(\%)$\downarrow$ & AUC(\%)$\uparrow$& HTER(\%)$\downarrow$ & AUC(\%)$\uparrow$ \\
        \midrule
        \multicolumn{11}{c}{\textbf{Uni-modal DG (Concat + $1\!\times\!1$ Conv)}}\\
        \midrule
        SSDG \cite{jia2020single} &26.09&82.03&28.50&75.91&41.82&60.56&40.48&62.31&37.32&68.25\\
        SSAN \cite{wang2022domain} &17.73&91.69&27.94&79.04&34.49&68.85&36.43&69.29&35.34&70.98\\
        SA-FAS \cite{sun2023rethinking} &21.37&87.65&23.22&84.49&35.10&70.86&35.38&69.71&28.77&78.18\\
        IADG \cite{zhou2023instance} &27.02&86.50&23.04&83.11&32.06&73.83&39.24&63.68&39.83&62.95\\
        FLIP \cite{liu2023ma}
        &13.19&93.79&11.73&94.93&17.39&90.63&22.14&83.95&16.11&90.83\\
        \midrule
        \multicolumn{11}{c}{\textbf{Multi-modal FAS}}\\
        \midrule
        ViT  \cite{dosovitskiy2020image} &20.88&84.77&44.05&57.94&33.58&71.80&42.15&56.45&36.60&68.12\\
        AMA \cite{yu2024rethinking} &17.56&88.74&27.50&80.00&21.18&85.51&47.48&55.56&27.47&79.85\\
        VP-FAS \cite{yu2023visual} &16.26&91.22&24.42&81.07&21.76&85.46&39.35&66.55&29.82&76.62\\
        ViTAF \cite{huang2022adaptive} &20.58&85.82&29.16&77.80&30.75&73.03&39.75&63.44&33.89&71.54\\
        MM-CDCN \cite{yu2020multi} &38.92&65.39&42.93&59.79&41.38&61.51&48.14&53.71&46.81&53.43\\
        CMFL \cite{george2021cross} &18.22&88.82&31.20&75.66&26.68&80.85&36.93&66.82&31.01&75.07\\
        MMDG \cite{lin2024suppress} &12.79&93.83&15.32&92.86&18.95&88.64&29.93&76.52&19.24&87.96\\
        DADM \cite{yang2025dadm} &11.71&94.89&6.92&97.66&19.03&88.22&16.87&91.08&13.63&92.96\\
        \midrule
        CLIP \cite{radford2021learning}
        &14.55&90.47&18.17&90.02&24.13&83.15&38.33&65.71&24.63&83.00\\
        \textbf{MMDA (Ours)}&\textbf{1.22}&\textbf{99.99}&\textbf{4.21}&\textbf{98.62}&\textbf{4.34}&\textbf{98.58}&\textbf{6.25}&\textbf{98.18}&\textbf{4.00}&\textbf{98.94}\\
        \bottomrule[1.3pt]
    \end{tabular}
    }
    \label{P1}
    \vspace{-2.0em}
\end{table*}

\begin{table*}[t]
    \centering
    \caption{Cross-dataset testing results under the missing modalities scenarios (Protocol 2) among CASIA-CeFA (C), PADISI (P), CASIA-SURF (S), and WMCA (W). Best results are marked in \textbf{bold}.} 
    \vspace{-0.8em}
    \tabcolsep=2mm
    \resizebox{1.00\textwidth}{!}{
    \begin{tabular}{l|cc|cc|cc|cc} 
        \toprule[1.3pt]
        \multirow{2}{*}{\textbf{Method}}& \multicolumn{2}{c|}{\textbf{Missing D}}&\multicolumn{2}{c|}{\textbf{Missing I}}&\multicolumn{2}{c|}{\textbf{Missing D \& I}}&\multicolumn{2}{c}{\textbf{Average}}\\
        \cmidrule(r){2-3} \cmidrule(r){4-5} \cmidrule(r){6-7} \cmidrule(r){8-9}
        & HTER(\%)$\downarrow$ & AUC(\%)$\uparrow$ & HTER(\%)$\downarrow$ & AUC(\%)$\uparrow$& HTER(\%)$\downarrow$ & AUC(\%)$\uparrow$ & HTER(\%)$\downarrow$ & AUC(\%)$\uparrow$\\
        \midrule
        \multicolumn{9}{c}{\textbf{Uni-modal DG (Concat + $1\!\times\!1$ Conv)}}\\
        \midrule
        SSDG \cite{jia2020single} &38.92&65.45&37.64&66.57&39.18&65.22&38.58&65.75\\
        SSAN \cite{wang2022domain} &36.77&69.21&41.20&61.92&33.52&73.38&37.16&68.17\\
        SA-FAS \cite{sun2023rethinking} &36.30&69.07&39.80&62.69&33.08&74.29&36.40&68.68\\
        IADG \cite{zhou2023instance} &40.72&58.72&42.17&61.83&37.50&66.90&40.13&62.49\\
        FLIP \cite{liu2023ma}&23.66&83.90&24.06&84.04&27.07&79.79&27.93&79.44\\
        \midrule
        \multicolumn{9}{c}{\textbf{Multi-modal FAS}}\\
        \midrule
        ViT \cite{dosovitskiy2020image}&40.04&64.69&36.77&68.19&36.20&69.02&37.67&67.30\\
        AMA \cite{yu2024rethinking} &29.25&77.70&32.30&74.06&31.48&75.82&31.01&75.86\\
        VP-FAS \cite{yu2023visual} &29.13&78.27&29.63&77.51&30.47&76.31&29.74&77.36\\
        ViTAF \cite{huang2022adaptive} &34.99&73.22&35.88&69.40&35.89&69.61&35.59&70.64\\
        MM-CDCN \cite{yu2020multi} &44.90&55.35&43.60&58.38&44.54&55.08&44.35&56.27\\
        CMFL \cite{george2021cross} &31.37&74.62&30.55&75.42&31.89&74.29&31.27&74.78\\
        MMDG \cite{lin2024suppress} &24.89&82.39&23.39&83.82&25.26&81.86&24.51&82.69\\
        DADM \cite{yang2025dadm} &21.56&85.17&20.82&85.28&22.61&84.04&21.66&84.83\\
        \midrule
        CLIP \cite{radford2021learning}&28.07&77.00&29.10&77.04&32.58&73.36&33.83&71.11\\
        \textbf{MMDA (Ours)}&\textbf{11.10}&\textbf{93.97}&\textbf{5.98}&\textbf{98.30}&\textbf{13.36}&\textbf{93.74}&\textbf{10.14}&\textbf{95.33}\\
        \bottomrule[1.3pt]
    \end{tabular}
    }
    \vspace{-1.5em}
    \label{P2}
\end{table*}
\begin{table}[t]
\centering

\begin{minipage}[t]{0.56\textwidth}
\centering
\caption{Cross-dataset testing results under the limited source domain scenarios (Protocol 3) among CASIA-CeFA (C), PADISI (P), CASIA-SURF (S), and WMCA (W). The best results are in \textbf{bold}.}
\label{P3}
\vspace{-1.2em}
\tabcolsep=0.2cm
\renewcommand{\arraystretch}{1.12}
\resizebox{1.00\linewidth}{!}{
\begin{tabular}{c|cc|cc}
    \toprule[1.3pt]
    \multirow{2}{*}{\textbf{Method}}  & \multicolumn{2}{c|}{\textbf{CW$\rightarrow$PS}}&\multicolumn{2}{c}{\textbf{PS$\rightarrow$CW}} \\
    \cmidrule(r){2-3} \cmidrule(r){4-5}
    & HTER(\%)$\downarrow$ & AUC(\%)$\uparrow$ & HTER(\%)$\downarrow$ & AUC(\%)$\uparrow$\\
    \midrule
    \multicolumn{5}{c}{\textbf{Uni-modal DG (Concat + $1\!\times\!1$ Conv)}}\\
    \midrule
    SSDG \cite{jia2020single}&25.34&80.17&46.98&54.29\\
    SSAN \cite{wang2022domain}&26.55&80.06&39.10&67.19\\
    SA-FAS \cite{sun2023rethinking}&25.20&81.06&36.59&70.03\\
    IADG \cite{zhou2023instance}&22.82&83.85&39.70&63.46\\
    FLIP \cite{liu2023ma}&15.92&92.38&23.85&83.46\\
    \midrule
    \multicolumn{5}{c}{\textbf{Multi-modal FAS}}\\
    \midrule
    ViT \cite{dosovitskiy2020image}&42.66&57.80&42.75&60.41\\
    AMA \cite{yu2024rethinking}&29.25&76.89&38.06&67.64\\
    VP-FAS \cite{yu2023visual}&25.90&81.79&44.37&60.83\\
    ViTAF \cite{huang2022adaptive}&29.64&77.36&39.93&61.31\\
    MM-CDCN \cite{yu2020multi}&29.28&76.88&47.00&51.94\\
    CMFL \cite{george2021cross}&31.86&72.75&39.43&63.17\\
    MMDG \cite{lin2024suppress}&20.12&88.24&36.60&70.35\\
    DADM \cite{yang2025dadm}&12.61&93.81&20.40&89.51\\
    \midrule
    CLIP \cite{radford2021learning}&19.36&90.57&29.98&79.22\\
    \textbf{MMDA (Ours)}&\textbf{7.52}&\textbf{96.84}&\textbf{6.30}&\textbf{98.35}\\
    \bottomrule[1.3pt]
\end{tabular}}
\end{minipage}
\hfill
\begin{minipage}[t]{0.42\textwidth}
\centering

\captionof{table}{Ablation results on MD2A. The best results are in \textbf{bold}.}
\vspace{-0.8em}
\label{Ablation_MHD2A}
\tabcolsep=0.2cm
\resizebox{1.00\linewidth}{!}{
\begin{tabular}{c|c|c}
    \toprule[1.3pt]
    \textbf{Method}&HTER(\%)$\downarrow$&AUC(\%)$\uparrow$ \\
    \midrule
    Dense Adaptor&23.26&84.92\\
    Dense Adaptor (w/ MHSA)&25.85&82.95\\
    Dense Adaptor (w/ DA)&16.49&92.05\\
    \textbf{Dense Adaptor (w/ MD2A)}&\textbf{13.47}&\textbf{94.20}\\
    \midrule
    MoE Adaptor&22.92&85.84\\
    MoE Adaptor (w/ MHSA)&12.83&93.25\\
    MoE Adaptor (w/ DA)&12.72&93.89\\
    \textbf{MoE Adaptor (w/ MD2A)}&\textbf{9.70}&\textbf{95.23}\\
    \bottomrule[1.3pt]
\end{tabular}}

\vspace{0.2em}

\captionof{table}{Ablation results on RS2. The best results are in \textbf{bold}.}
\label{Ablation_DS2}
\vspace{-0.8em}
\tabcolsep=0.2cm
\resizebox{1.00\linewidth}{!}{
\begin{tabular}{c|c|c}
    \toprule[1.3pt]
    \textbf{Method}&HTER(\%)$\downarrow$&AUC(\%)$\uparrow$ \\
    \midrule
    Vanilla Alignment&9.70&95.23\\
    Smooth Alignment&9.17&96.32\\
    \textbf{RS2 Alignment}&\textbf{8.88}&\textbf{97.20}\\
    \bottomrule[1.3pt]
\end{tabular}}

\vspace{0.2em}

\captionof{table}{Pairing-swap invariance analysis on MD2A.}
\label{tab:pairing_swap_invariance_compact}
\vspace{-0.8em}
\tabcolsep=0.2cm
\renewcommand{\arraystretch}{1}
\resizebox{1.00\linewidth}{!}{
\begin{tabular}{lcc}
    \toprule[1.3pt]
    Pairing strategy & \(\mathrm{Var}[R]\downarrow\) & \(\mathrm{Var}[z]\downarrow\) \\
    \midrule
    Same-domain  & \(4.31\times 10^{-7}\) & \(1.672\times 10^{-3}\) \\
    Random       & \(7.48\times 10^{-7}\) & \(1.662\times 10^{-3}\) \\
    Cross-domain & \(6.94\times 10^{-7}\) & \(1.728\times 10^{-3}\) \\
    \bottomrule[1.3pt]
\end{tabular}}

\end{minipage}
\vspace{-2.0em}
\end{table}

\vspace{-0.3em}
\subsection{Cross-domain Testing}
\vspace{-0.3em}
\noindent\textbf{Protocol 1: Complete modality scenario.}\quad
Protocol~1 evaluates cross-domain generalization when all modalities are available in both training and testing. 
As shown in Table~\ref{P1}, MMDA consistently outperforms both unimodal DG baselines and multimodal FAS methods on all four LOO splits. 
Compared with the strongest prior multimodal DG method DADM~\cite{yang2025dadm}, MMDA reduces the average HTER from 13.63\% to 4.00\% and increases the average AUC from 92.96\% to 98.94\% (+5.98\%). 
All four sub-protocols achieve HTER $\le$ 6.25\% and AUC $\ge$ 98.18\%, indicating that the learned multimodal representations transfer well to unseen domains. 
These results support our design choice of first shrinking domain and modality gaps via MD2A and then aligning multimodal features in a CLIP-based representation space using RS2 and U-DSA, rather than relearning a new fused decision boundary from scratch.

\noindent\textbf{Protocol 2: Missing modality scenario at test time.}\quad
Protocol~2 evaluates robustness when some modalities are missing during testing. 
Table~\ref{P2} shows that MMDA again achieves the best performance in all three missing-modality settings. 
On average, MMDA reduces HTER from 21.66\% (DADM) to 10.14\% and improves AUC from 84.83\% to 95.33\%, without any modality-specific dropout or retraining for missing modalities. 
When IR is missing, MMDA still obtains 5.98\% HTER and 98.30\% AUC, very close to the complete-modality case, suggesting that the purified RGB and DEPTH features after MD2A and their soft alignment via RS2 can compensate for the loss of IR. 
In contrast, removing DEPTH leads to a larger degradation (11.10\% HTER and 93.97\% AUC), highlighting that DEPTH provides complementary geometric information that is harder to recover from RGB and IR alone, even under the proposed purify then guide scheme.

\noindent\textbf{Protocol 3: Limited source domain scenario.}\quad
Protocol~3 restricts the number of source domains to two, simulating limited-source training. 
In Table~\ref{P3}, MMDA performs strongly in both CW$\rightarrow$PS and PS$\rightarrow$CW. 
In the CW$\rightarrow$PS setting, MMDA improves HTER from 12.61\% (DADM) to 7.52\% and AUC from 93.81\% to 96.84\%. 
In the more challenging PS$\rightarrow$CW setting, where WMCA shows substantial distribution shift, MMDA reduces HTER from 20.40\% to 6.30\% and raises AUC from 89.51\% to 98.35\%. 
The improvements under limited-source training further confirm that explicitly purifying multimodal features and leveraging CLIP-based soft alignment allows MMDA to retain generalizable cues even when domain coverage is scarce.

\vspace{-0.8em}
\subsection{Ablation Study}
\vspace{-0.5em}

\noindent\textbf{Effectiveness of MD2A.}\quad
Table~\ref{Ablation_MHD2A} evaluates MD2A under two variants on the PS$\rightarrow$CW split. 
Starting from the Dense adapter baseline (23.26\% HTER / 84.92\% AUC), adding a standard multi-head self-attention (MHSA) even hurts performance. 
Replacing MHSA with Differential Attention (DA)~\cite{ye2024differential} yields clear gains (16.49\% / 92.05\%), and our MD2A further reduces HTER to 13.47\% and raises AUC to 94.20\%. 
A similar trend holds for the MoE adapter: MD2A improves over both MHSA and DA, achieving 9.70\% HTER and 95.23\% AUC. 
These results indicate that reformulating DA with same-domain, cross-sample pairing indeed helps shrink domain and modality gaps in the fused representation, going beyond sample-wise DA and better realizing the ``purify'' step in our framework.

\noindent\textbf{Effectiveness of RS2 Alignment.}\quad
Table~\ref{Ablation_DS2} studies different alignment strategies in the CLIP representation space. 
“Vanilla Alignment” directly penalizes distances between visual features and their corresponding text prompts, obtaining 9.70\% HTER and 95.23\% AUC. 
Introducing “Smooth Alignment” improves both metrics (9.17\% / 96.32\%), suggesting that overly hard alignment targets are suboptimal. 
Our full RS2 loss, which combines smoothed supervision with nearest-text subspace alignment and a shared classifier on both visual and text embeddings, further reduces HTER to 8.88\% and increases AUC to 97.20\%. 
This confirms that explicitly shaping the geometry of the CLIP space through soft subspace alignment, rather than merely updating model weights or aligning features with a single prompt, yields consistent improvements in multimodal DG FAS and effectively realizes the “guide” step.

\begin{figure}[t]
\centering
\begin{minipage}[t]{0.49\textwidth}
    \centering
    \includegraphics[width=\linewidth]{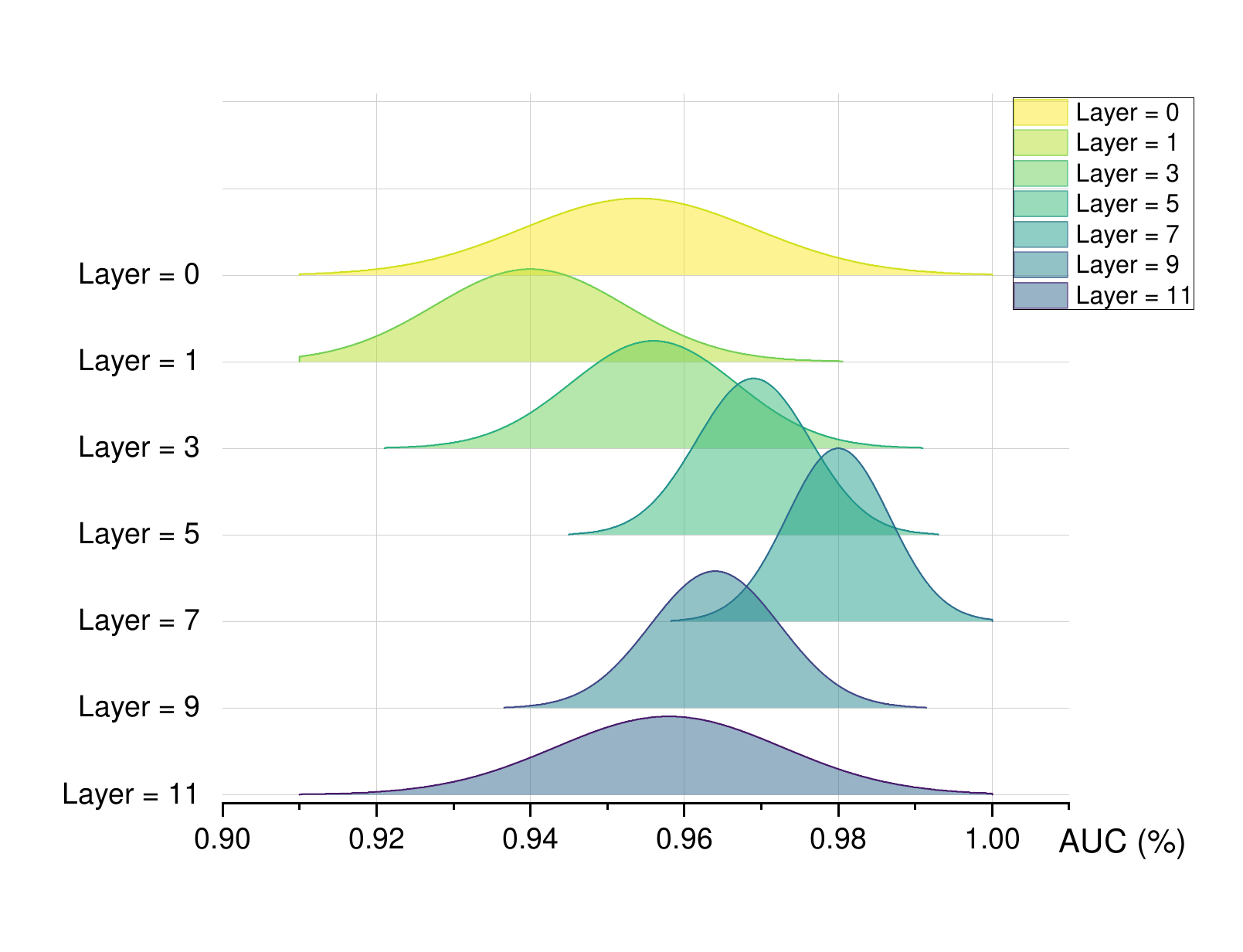}
    \vspace{-3.0em}
    \caption{AUC statistics of the U-DSA module across various caption groups at different depths. The height of each bar represents the number of captions achieving the specified AUC.}
    \label{fig:ablation_udsa}
\end{minipage}\hfill
\begin{minipage}[t]{0.49\textwidth}
    \centering
    \includegraphics[width=\linewidth]{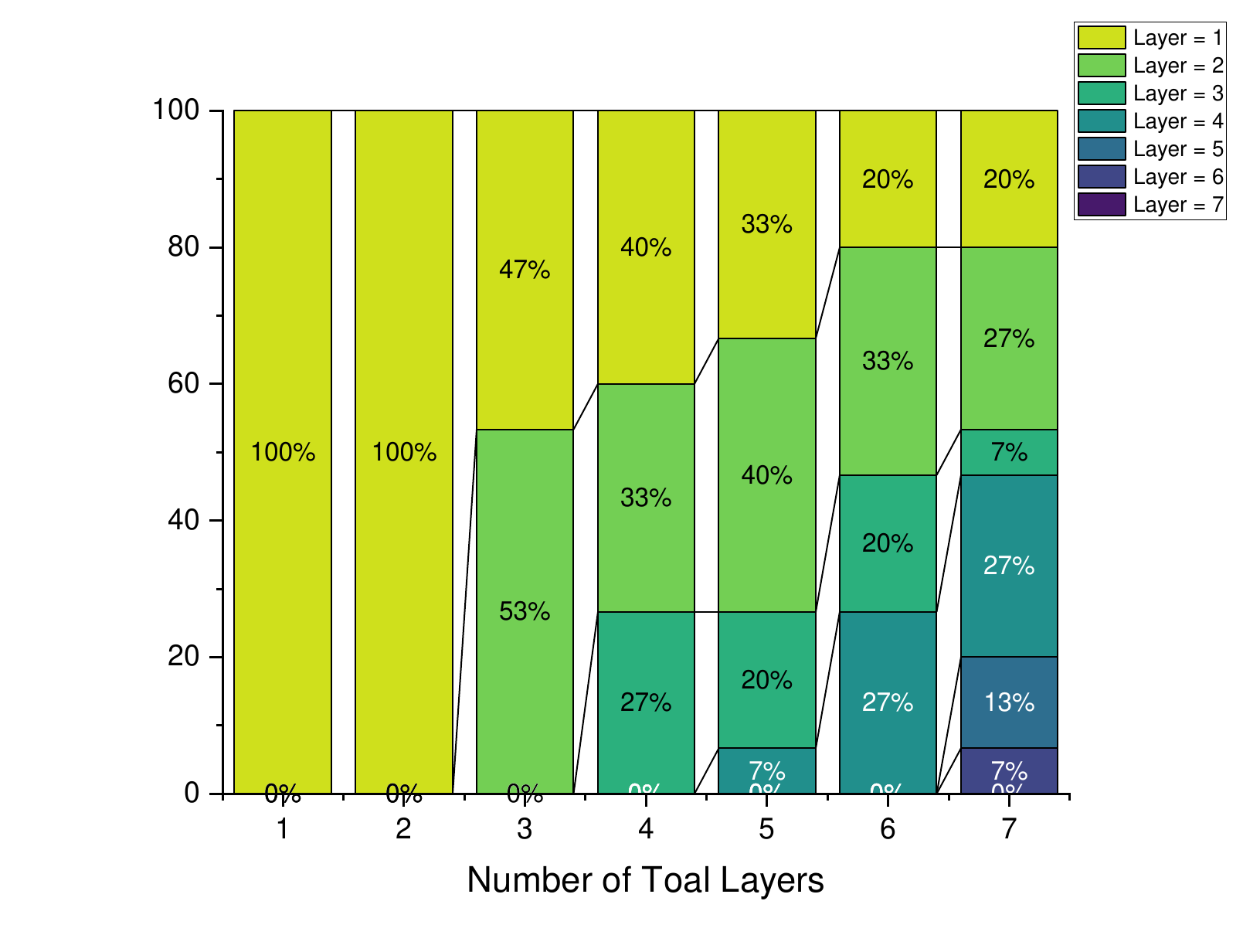}
    \vspace{-3.0em}
    \caption{Statistics of the layers achieving the best alignment effects in representation spaces constructed by U-DSA under different total depths (1 to 7 layers).}
    \label{Ablation EQ}
\end{minipage}
\vspace{-2.0em}
\end{figure}

\noindent\textbf{Effectiveness of U-DSA.}\quad
We next analyze the impact of U-DSA depth and its dual-space design. 
Fig.~\ref{fig:ablation_udsa} summarizes, for different caption sets (representation spaces), how many caption groups achieve a given AUC at each U-DSA depth. 
The distribution shifts towards higher AUC values as the depth increases up to 7 layers, indicating that deeper adaptation modules help capture richer multimodal patterns when combined with MD2A and RS2. 
A complementary view is given in Fig.~\ref{Ablation EQ}: for total depths from 1 to 7, the layers that yield the best performance are predominantly shallow or mid-depth, and the deepest layer never achieves the best AUC. 
This matches our design intuition for U-DSA: deep adaptation is useful, but the final classifier should operate on a remapped representation closer to the shallower CLIP space, rather than on the deepest features alone.

\noindent\textbf{Denoising mechanism analysis.}
We interpret MD2A as estimating and cancelling domain-/modality-correlated common-mode nuisance under same-domain conditioning, thereby purifying the representation while preserving spoof discriminative cues. To validate this mechanism, we propose a pairing-swap invariance test (Table~\ref{tab:pairing_swap_invariance_compact}). Specifically, for each sample $x$, the main branch outputs $F_{\mathrm{main}}(x)$. Given a paired sample $\tilde{x}$, the differential branch predicts a residual $R(x,\tilde{x})$, and we form the denoised feature as
\begin{equation}
\small
F_{\mathrm{de}}(x,\tilde{x}) = F_{\mathrm{main}}(x) - R(x,\tilde{x}).
\end{equation}
We keep $x$ fixed and only swap $\tilde{x}$ with three pairing strategies (\emph{same-domain} / \emph{random} / \emph{cross-domain}), repeating pairing 10 times for each $x$.
We then measure: (i) \textbf{residual stability} via within-swap $\mathrm{Var}[R]$; (ii) \textbf{decision stability} via within-swap $\mathrm{Var}[z]$, where $z$ is the spoof logit computed from $F_{\mathrm{de}}$; and (iii) \textbf{domain leakage} via HSIC between features and domain labels (lower indicates weaker dependence).
As shown in Table~\ref{tab:pairing_swap_invariance_compact}, \emph{same-domain} pairing yields the lowest $\mathrm{Var}[R]$ and exhibits a clear mean-residual drift when switching to \emph{cross-domain} pairing (drift magnitude: $9.55\!\times\!10^{-7}$).
Meanwhile, domain dependence drops substantially after subtraction (HSIC: $1.596\!\rightarrow\!0.269$, drop $\!=\!1.327$), while $\mathrm{Var}[z]$ remains small. Together, these results support that MD2A performs common-mode noise suppression rather than arbitrary semantic subtraction, without destabilizing spoof decisions.

\vspace{-0.5em}
\subsection{Visualization and Analysis}
\vspace{-0.5em}

\begin{figure}[t]
\centering
\begin{minipage}[t]{0.49\textwidth}
    \centering
    \raisebox{2em}{\includegraphics[width=\linewidth,trim=14em 6em 6em 0,clip]{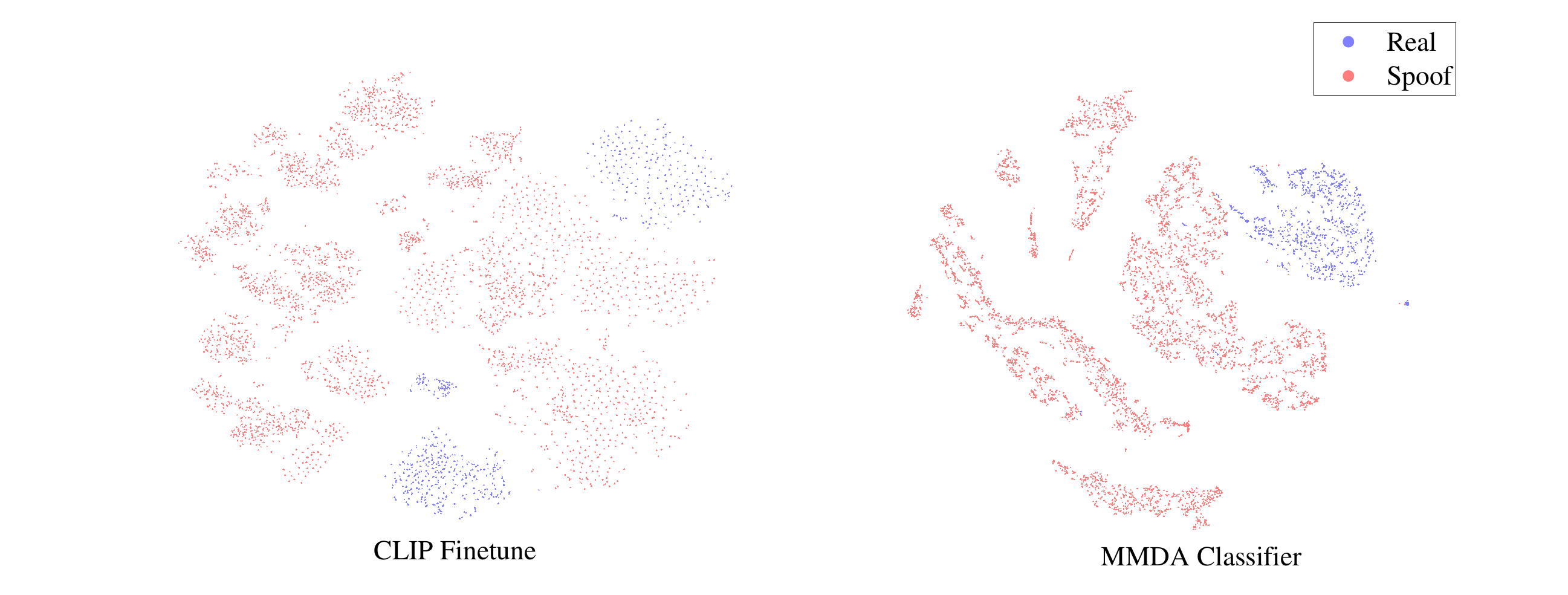}}
    \vspace{-3.35em}
    \caption{t-SNE visualization of the fine-tuned CLIP baseline (left) and the classifier part of MMDA (right).}
    \label{fig:tsne_compare}
\end{minipage}\hfill
\begin{minipage}[t]{0.49\textwidth}
    \centering
    \includegraphics[width=0.8\linewidth]{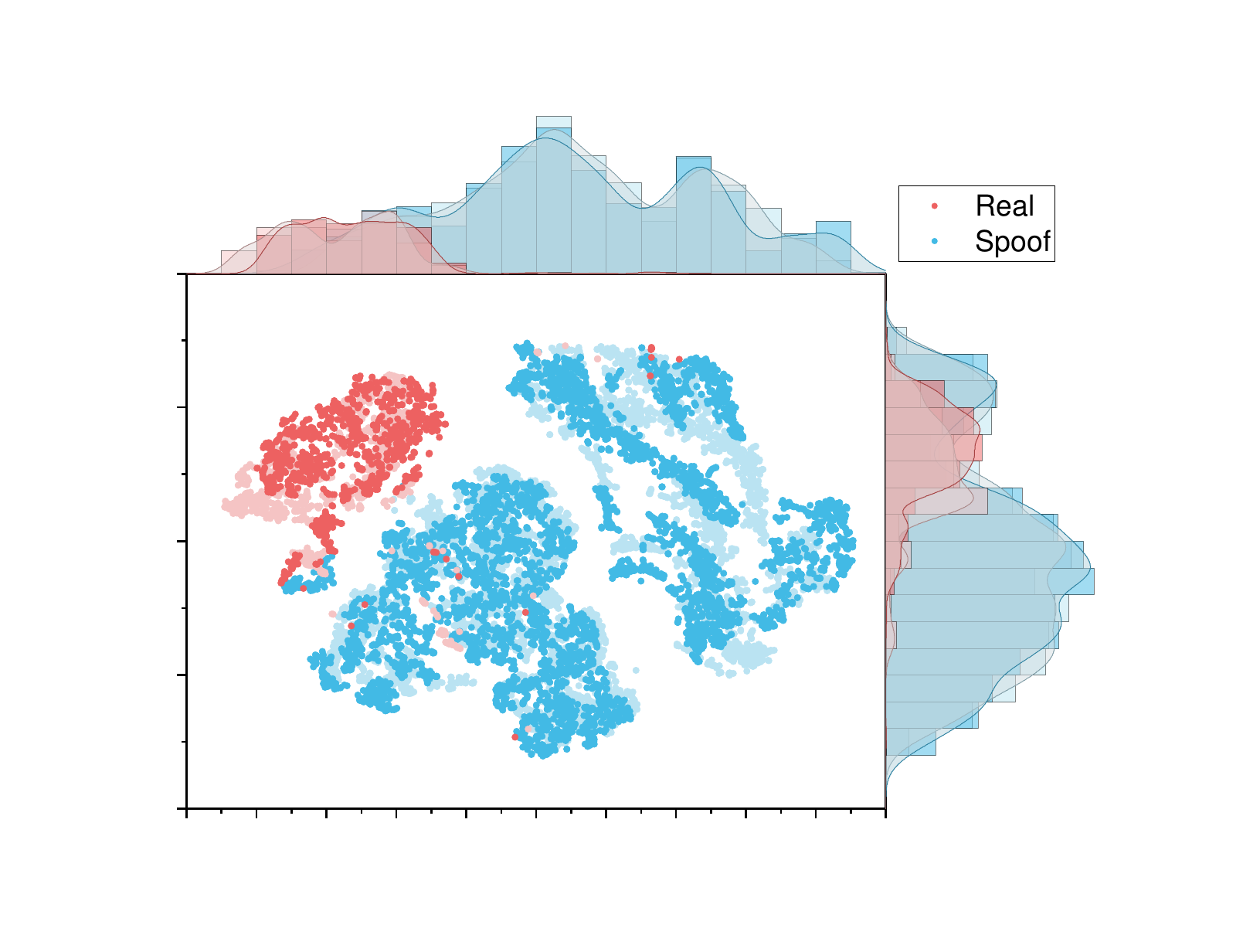}
    \vspace{-2.0em}
    \caption{t-SNE visualization of representations before entering U-DSA (lighter color) and after the deepest layer (darker color).}
    \label{fig:tsne_depth}
\end{minipage}
\vspace{-2.0em}
\end{figure}

The t-SNE plots in Fig.~\ref{fig:tsne_compare} compare the fine-tuned CLIP baseline with MMDA. 
For CLIP, real and spoof samples from different domains form several overlapping clusters, and cross-domain mixing within each class is limited. 
In contrast, MMDA produces a much clearer separation between real and spoof, with samples from different domains better mixed inside each class cluster, consistent with the goal of first reducing domain and modality gaps and then aligning to a stable CLIP-based real/spoof separation.

Fig.~\ref{fig:tsne_depth} visualizes how representations evolve along U-DSA. 
From the input (lighter points) to the output after deep adaptation and remapping (darker points), samples gradually concentrate into tighter, more compact clusters. 
This indicates that U-DSA refines features while keeping them close to a stable decision structure, rather than drifting away from the pre-trained CLIP geometry.

\vspace{-1.0em}
\section{Conclusion}
\vspace{-1.0em}
We presented \textbf{M}ultimodal \textbf{D}enoising and \textbf{A}lignment (MMDA), a CLIP-based framework that tackles multimodal domain-generalized face anti-spoofing from a ``purify then guide'' perspective: MD2A first \emph{purifies} fused RGB/IR/DEPTH features by suppressing domain-/modality-specific artifacts, and RS2 together with U-DSA then \emph{softly guides} the purified representations within a CLIP-based real/spoof space without destroying its geometry. Extensive experiments on WMCA, CeFA, PADISI, and SURF under complete-modality, missing-modality, and limited-source protocols show substantial and consistent gains over prior unimodal DG and multimodal FAS methods. Limitations remain, e.g., mixture-of-experts layers can be optimization-sensitive in pre-trained spaces. Future work will focus on more stable and adaptive dual-space adaptation and richer, more inclusive caption sets to further enhance generalization across unseen domains, sensors, and attack types.


\section*{Acknowledgements}
This work was supported in part by the National Natural Science Foundation of China (Grant Nos. 62576216 and 62576076), the Guangdong Basic and Applied Basic Research Foundation (Grant No. 2023A1515140037), the Guangdong Provincial Key Laboratory (Grant No. 2023B1212060076), the CCF--Tencent Rhino-Bird Open Research Fund, and the Guangdong Research Team for Communication and Sensing Integrated with Intelligent Computing (Project No. 2024KCXTD047). Computational resources were provided by the SongShan Lake HPC Center (SSL-HPC) at Great Bay University.

%
%
\bibliographystyle{splncs04}
\bibliography{main}

\end{document}